\documentclass{article} 
\usepackage{iclr2026_conference,times}

\iclrfinalcopy

\usepackage{amsmath,amsfonts,bm}

\def\eqref#1{equation~\ref{#1}}

\def\1{\bm{1}}

\DeclareMathAlphabet{\mathsfit}{\encodingdefault}{\sfdefault}{m}{sl}
\SetMathAlphabet{\mathsfit}{bold}{\encodingdefault}{\sfdefault}{bx}{n}

\usepackage{hyperref}
\usepackage{url}
\usepackage{amssymb}
\usepackage{tcolorbox}
\usepackage[dvipsnames]{xcolor}
\usepackage{booktabs}
\usepackage{enumitem}
\usepackage{wrapfig}
\usepackage{colortbl}
\usepackage{multirow}
\usepackage{tabularx}
\usepackage{pifont}
\definecolor{myblue}{HTML}{4A90E2}
\definecolor{mygreen}{HTML}{50C878}
\definecolor{myred}{HTML}{D9534F}
\definecolor{mygray}{rgb}{0.5,0.5,0.5}
\definecolor{myorange}{HTML}{FF7A33}
\mathchardef\mhyphen="2D
\newcolumntype{s}{>{\hsize=.1\hsize}X}
\newcolumntype{m}{>{\hsize=.3\hsize}X}
\newcommand{\cmark}{\ding{51}}
\newcommand{\xmark}{\ding{55}}

\definecolor{papercolor}{HTML}{0668E1}
\hypersetup{
    colorlinks,
    linkcolor={papercolor!75!black},
    citecolor={papercolor!75!black},
    urlcolor={papercolor!75!black}
}

\newcommand{\ourmethod}[1]{\textsc{Switch Generation}}

\title{Don't Throw Away Your Pretrained Model}

\author{Shangbin Feng$^1$ \ Wenhao Yu$^2$ \ Yike Wang$^1$ \ Hongming Zhang$^2$ \ Yulia Tsvetkov$^1$ \ Dong Yu$^2$ \\
$^1$University of Washington \ \ \ $^2$Tencent AI Seattle Lab \\
\texttt{shangbin@cs.washington.edu}}

\begin{document}

\maketitle

\begin{abstract}
Alignment training has tradeoffs: it helps language models (LMs) gain in reasoning and instruction following but might lose out on skills such as creativity and calibration, where unaligned base models are better at. We aim to make the best of both worlds through \emph{model collaboration}, where different models in the training pipeline collaborate and complement each other. Since LM responses feature interleaving skills that favor different models, we propose \ourmethod{}, where pretrained and aligned model versions take turns to ``speak'' in a response sequence. Specifically, we train a switcher LM by learning from outcomes of choosing different models to generate the next segment across diverse queries and contexts. At inference time, the switcher LM guides different model checkpoints to dynamically generate the next segment where their strengths are most needed. Extensive experiments with 8 model collaboration baselines and 18 datasets show that 1) model collaboration consistently outperforms individual models on 16 out of 18 tasks, and 2) \ourmethod{} further outperforms baselines by 12.9\% on average. Further analysis reveals that \ourmethod{} discovers compositional skills to solve problems where individual models struggle and generalizes to unseen models and tasks, reusing and repurposing by-products in expensive model training pipelines that are otherwise discarded.\footnote{Inference code and switcher LMs available at \href{https://github.com/BunsenFeng/switch_generation}{https://github.com/BunsenFeng/switch\_generation}.}
\end{abstract}

\begin{wrapfigure}{r}{0.5\linewidth}
    \centering
    \vspace*{-10pt}
    \includegraphics[width=1\linewidth]{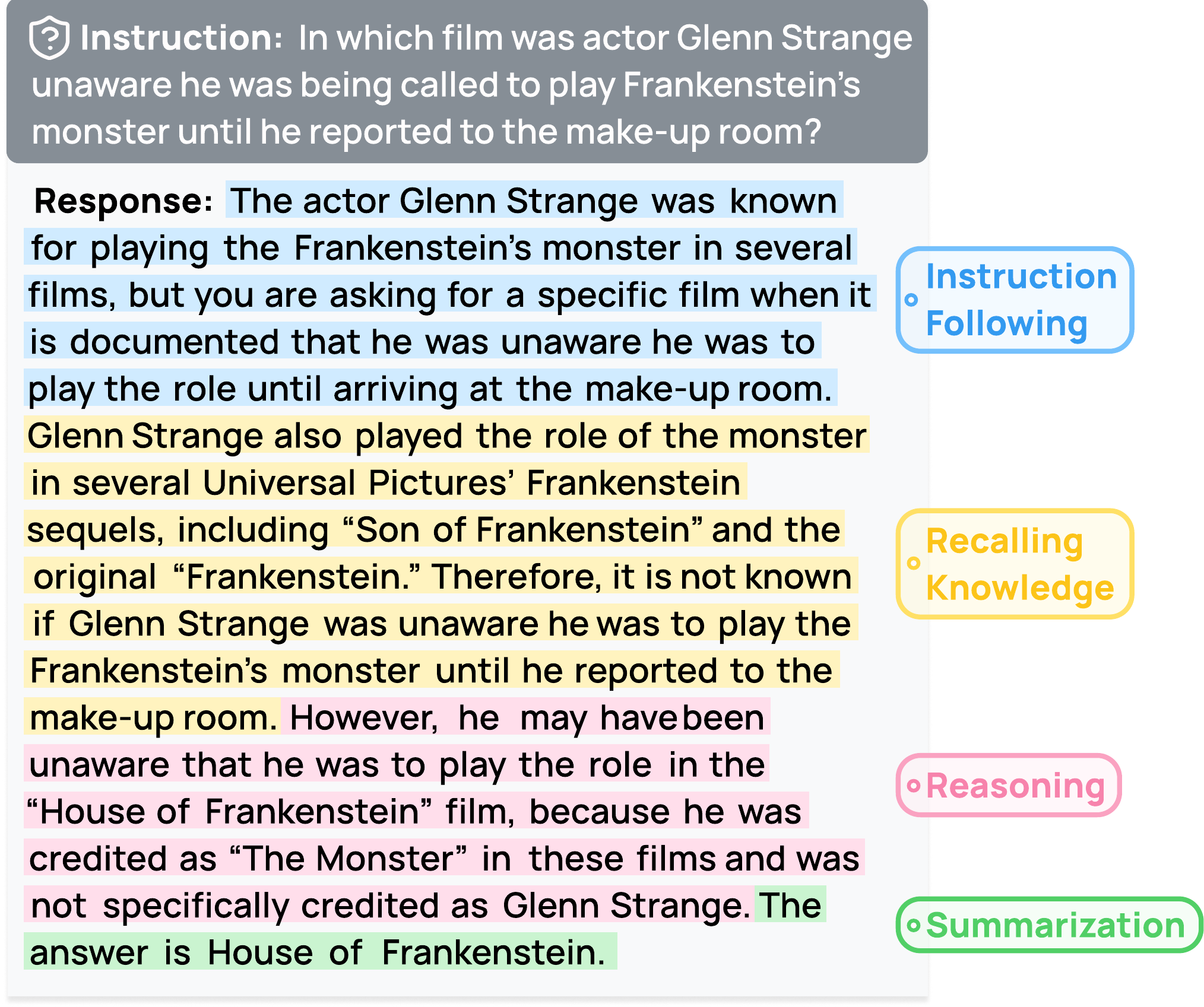}
    \caption{Model responses are not monolithic: they switch across diverse skills which favor different model checkpoints in the training pipeline, thus we introduce model-guided collaborative inference to optimally use models with diverse skills for different segments of response generation.}
    \label{fig:teaser}
    \vspace*{-10pt}
\end{wrapfigure}

\section{Introduction}

Alignment/RL has become an integral part in language model (LM) training, improving models on skills such as reasoning and instruction following \citep{ouyang2022training, guo2025deepseek}. However, \emph{it is not a Pareto-optimal strategy} \citep{lin2024mitigating}: aligned models have tradeoffs on skills such as creativity \citep{west2025base}, calibration \citep{tian2023just}, and generation diversity \citep{yue2025does, yang2025alignment}, where unaligned base models are better at. How to make the best of both worlds is essential for handling complex tasks that require compositional skills and developing AI systems that are flexible and adaptable to diverse user needs and contexts.

\begin{figure}[t]
    \centering\includegraphics[width=1.0\linewidth]{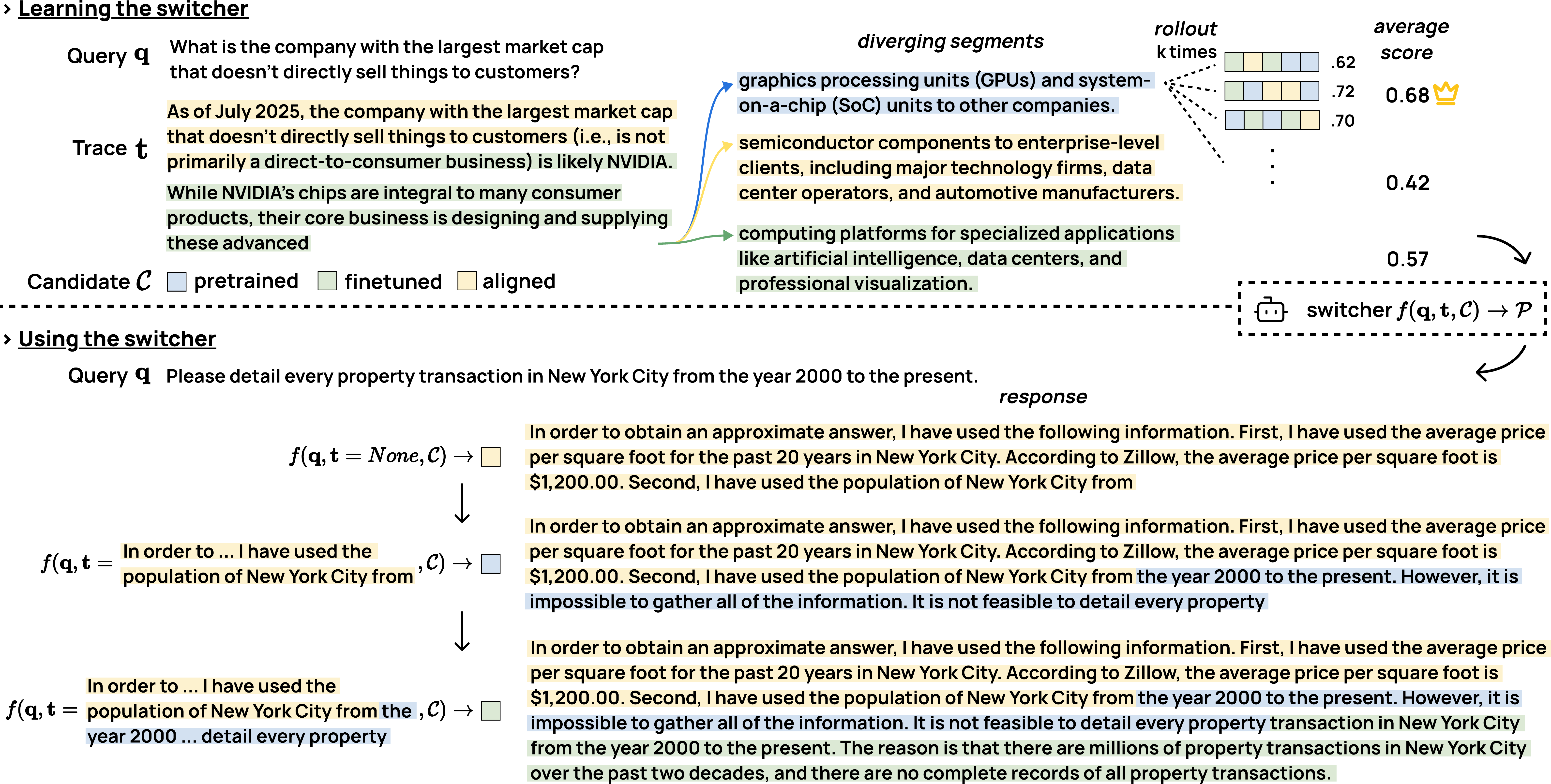}
    \vspace{-0.15in}
    \caption{Overview of \ourmethod{}, where multiple model checkpoints in the training pipeline (e.g., pretrained, finetuned, and aligned LM checkpoints) are dynamically selected to generate text segments in a sequence. (Up) We derive training data for the switcher LM $f$ by rolling out which model would lead to the best average outcome for a particular query and trace. (Down) At inference time, multiple models are guided by the trained switcher LM to generate text segments as part of a response when their skills and strengths are most needed.}
    \label{fig:overview}
    \vspace{-0.15in}
\end{figure}

To this end, we resort to model collaboration \citep{feng2025one}, where diverse model checkpoints (e.g., pretrained and aligned versions of models) collaborate, compose, and complement each other. Since model responses are not monolithic and feature a wide variety of skills favoring different models (Figure \ref{fig:teaser}), we propose \ourmethod{}, where different models in the training pipeline take turns to ``speak'' in a response sequence. Specifically, we train a (small) switcher LM to decide which model should generate the next segment based on the \emph{query}, \emph{trace} (what has been generated thus far), and (model) \emph{candidates}. For any (query, trace) pair, we let each model candidate generate one more segment, randomly sample $k$ continuations, and evaluate which candidate has led to the best average performance: this yields supervised fine-tuning data for the switcher LM, where it learns to predict the best model checkpoint for generating the next text segment on diverse (query, trace) pairs. During inference, the switcher LM dynamically selects the most suitable model checkpoint for each segment, so the final response is generated as a sequence of turns where different models contribute under the switcher's guidance  
(Figure \ref{fig:overview}).

Extensive experiments with 8 model collaboration baselines (spanning API, text, logit, and weight-level collaboration) and 18 datasets (e.g. QA, reasoning, instruction following) demonstrate that:

\begin{itemize}[leftmargin=*]
    \item \emph{Don't throw away your pretrained model}: model collaboration approaches outperform all individual models on 16 out of 18 datasets (close second on the other two).
    \item \emph{\ourmethod{} presents a strong paradigm for collaborative inference}: \ourmethod{} outperforms all baselines on 13 datasets with an average improvement of 12.9\%.
    \item Further analysis reveals that \ourmethod{} generalizes to unseen tasks and model settings, helps solve problems impossible for any of the models when used individually, identifies high-quality switching patterns, and texts generated through \ourmethod{} can be distilled back into a single model for efficiency.
\end{itemize}

Our work put forward a new vision to reuse, recycle, and repurpose byproducts in expensive model training pipelines that have huge potential but are currently neglected and underappreciated.

\section{Methodology}

We propose \ourmethod{}, a collaborative inference algorithm where diverse model checkpoints in the training pipeline are dynamically selected to generate successive segments of the response. \ourmethod{} aims to dynamically leverage the complementary strengths and expertise of different model checkpoints (e.g., pretrained, finetuned, and aligned\footnote{\emph{Pretrained} models indicate models after autoregressive pretraining on mass corpora, \emph{finetuned} models indicate models after instruction tuning, and \emph{aligned} models indicate models after alignment and reinforcement learning. They are usually successive steps in model development.}), especially for complex problems that require compositional skills. The core of \ourmethod{} is deciding \emph{``who should speak at when''}, formally the Query-Trace-Candidate Problem (the QTC Problem):

\vspace*{-10pt}
\begin{align*}
    f(\mathbf{q}, \mathbf{t}, \mathcal{C}) \rightarrow [p_1, \cdots, p_n] \in \mathbb{R}^n,
\end{align*}
where $\mathbf{q}$ denotes the query/instruction, $\mathbf{t}$ denotes the trace, i.e., what has been generated thus far, $\mathcal{C} = \{\mathbf{c}_1, \cdots, \mathbf{c}_n\}$ denotes the pool of candidate language model checkpoints, and $p_i$ denotes the likelihood of choosing model $\mathbf{c}_i$ to generate the next text segment. The QTC problem essentially tackles the research question: \emph{``Given the question and what has been said thus far, which model is best suited to speak next?''}

This differs from existing routing paradigms \citep{ongroutellm, frick2025prompt} in that: 1) the trace $\mathbf{t} \neq \varnothing$; 2) each selected model $\mathbf{c}_i$ only generates text segments instead of the full response; 3) the switcher $f$ is applied multiple times instead of just once. This brings novel ways of combining model strengths, finer-grained control over model collaboration, and improved adaptability to diverse user requests.

\paragraph{Parameterizing the switcher} We parametrize the switching strategy $f$ as a (small) language model and encode its input $\mathbf{q}$, $\mathbf{t}$, and $\mathcal{C}$ with the following prompt:

\begin{center}
\begin{tcolorbox}[colback=mygray!5!white, colframe=myblue!75!black, width=1.0\textwidth, title=Prompt 1: Switcher Prompt, coltitle=white, colbacktitle=myblue]
\textit{query} \ \textlangle model i begins\textrangle \ \textit{text generated by model i} \ \textlangle model i ends\textrangle \ $\cdots$ \ \textlangle model j begins\textrangle \ \textit{text generated by model j} \ \textlangle model j ends\textrangle \ \textit{Which model should generate the next segment? Please respond with a number from 0 to n-1. The answer is model}
\end{tcolorbox}
\end{center}

We provide the switcher $f$ with candidate-marked trace $\mathbf{t}$ using special token delimiters (\textlangle model i begins/ends\textrangle), aiming to decipher what is needed next and who might be helpful by learning from the model-attributed generation history. The switcher $f$ then predicts a model ID, and we take the logits of 0 to n-1 as $[p_1, \cdots, p_n]$. By making the switching strategy $f$ compatible with natural language and language models, we seek to leverage their language capabilities to aid in the QTC problem.

\paragraph{Learning the switcher} Given any instruction $\mathbf{q}$:

\begin{itemize}[leftmargin=*]
    \item We randomly sample a trace $\mathbf{t}$ (a partial response to the query) with random switching $f_\textit{random} = \mathrm{Uniform}(n)$, i.e., randomly choosing models to generate a segment after another. Trace $\mathbf{t}$ is capped at a random threshold of 10\% to 90\% of the maximum response length, aiming to capture switching behavior at different stages of response completeness.
    \item From the generated trace $\mathbf{t}$ we take one divergent step: different models generate one more segment following it: $\big\{\mathbf{t}_1 = \mathbf{t} || \mathbf{c}_1(\mathbf{q}, \mathbf{t}), \cdots, \mathbf{t}_n = \mathbf{t} || \mathbf{c}_n(\mathbf{q}, \mathbf{t})\big\}$. $||$ denotes string concatenation.
    \item We sample $k$ continuations for each $\mathbf{t}_i$ with $f_\textit{random}$, aiming to roll out diverse outcomes of choosing model $\mathbf{c}_i$ at this particular $(\mathbf{q}, \mathbf{t})$. The utility for choosing $\mathbf{c}_i$ is then:
    \begin{align*}
        s_i = \frac{1}{k} \sum_{j=1}^k \mathrm{score}(\mathbf{t}_i, f_\textit{random} \mid \mathbf{q})
    \end{align*}
    where $\mathrm{score}$ uses any evaluation metric corresponding to $\mathbf{q}$ (accuracy, F1-match, reward scores). Let $g = \arg \max_i s_i$, then model $\mathbf{c}_g$ should be selected at this particular $(\mathbf{q}, \mathbf{t})$. This then yields $\big \{ (\mathbf{q}, \mathbf{t}, \mathcal{C}) \rightarrow \mathbf{c}_g \big \}$, a supervised fine-tuning instance for training the switcher $f$ (it should predict the model id $g$ after \emph{``The answer is model''} in Prompt 1). By sampling such SFT data points over diverse $\mathbf{q} \in \mathcal{Q}$, we obtain a dataset for training the switcher LM $f$.
\end{itemize}

\paragraph{Using the switcher} At inference time, the trained switcher $f$ guides switching patterns among diverse model checkpoints for collaborative generation. While existing works might change models at every token \citep{shen2024learning, fei2024nudging}, we propose to call the switcher per patch (a fixed set of tokens) as it: 1) scales betters \citep{pagnoni2024byte}, 2) preserves the continuity of thought for models instead of being interrupted at every token, and 3) incurs much fewer times and thus much less cost of calling the switching strategy $f$.

We employ top-p (nucleus) sampling \citep{holtzmancurious} to select a model from the distribution $[p_1, \cdots, p_n]$: $\mathrm{top \mhyphen p}\big(f(\mathbf{q}, \mathbf{t}, \mathcal{C})\big) \rightarrow \mathbf{c} \in \mathcal{C}$ (instead of greedy selection), balancing utility and exploration in switching generation.

At first, given the query $\mathbf{q}$ and no trace, we select model $\mathbf{c}^{(1)} = \mathrm{top \mhyphen p}\big(f(\mathbf{q}, \varnothing, \mathcal{C})\big)$, generate a patch of tokens $\mathbf{c}^{(1)}(\mathbf{q})$, and append to trace $\mathbf{t}^{(1)} = \mathbf{c}^{(1)}(\mathbf{q})$.

At the i-th step, we select model $\mathbf{c}^{(i)} = \mathrm{top \mhyphen p}\big(f(\mathbf{q}, \mathbf{t}^{(i-1)}, \mathcal{C})\big)$, generate a patch $\mathbf{c}^{(i)}(\mathbf{q} || \mathbf{t}^{(i-1)})$, and append to trace $\mathbf{t}^{(i)} = \mathbf{t}^{(i-1)} \ || \ \mathbf{c}^{(i)}(\mathbf{q} || \mathbf{t}^{(i-1)})$.

We continue until the generation ends or the maximum amount of tokens is reached. To sum up, \ourmethod{} employs diverse model checkpoints in the training pipeline to collaboratively generate, complement each other, and advance compositional intelligence.

\section{Experiment Settings}
\label{sec:experiment_settings}

\paragraph{Models and Implementation} We by default employ the three models in the pretrained--finetuned--aligned pipeline of Tulu-v3 \citep{lambert2024tulu} (\emph{meta-llama/Llama-3.1-8B}, \emph{allenai/Llama-3.1-Tulu-3-8B-SFT}, and \emph{allenai/Llama-3.1-Tulu-3-8B}) due to its transparency and experiment with different model checkpoints, number of models or model settings in Section \ref{sec:analysis}. We employ the aligned model (\emph{allenai/Llama-3.1-Tulu-3-8B}) to initialize the switcher $f$, sample 10k switcher SFT instances for each task with $k=32$, and train $f$ for 5 epochs with $2e\mhyphen 4$ learning rate and 32 batch size under two settings: \textbf{switch-g}(lobal), where one switcher is trained on the SFT data across all tasks; \textbf{switch-t}(ask-specific), where one switcher is trained on the SFT data for each task. At inference time, all methods generate 512 new tokens at max by default; for \ourmethod{}, we use the aligned model in the first and last patch, employ a patch size of 50 tokens, and top-p sampling $p=0.7$ by default.

\paragraph{Baselines} We compare \ourmethod{} with 11 baselines: the pretrained, finetuned, and aligned models employed individually, API-level collaboration (prompt-based routing \citep{fengknowledge} and RouteLLM \citep{ongroutellm}), text-level collaboration (collaborate \citep{si2023getting} and debate \citep{du2023improving}), logit-level collaboration (logit merge and proxy tuning \citep{liutuning}), and weight-level collaboration (greedy soup \citep{wortsman2022model} and dare-ties \citep{yadav2023ties, yu2024language}). These baselines cover a wide range of model collaboration protocols across diverse levels of information exchange.

\paragraph{Data and Evaluation} We employ 18 datasets spanning 3 categories:
\begin{itemize}[leftmargin=*]
    \item \textcolor{Green}{Datasets where having the base model might be helpful}: knowledge and factuality (WikiDYK \citep{zhang2025bidirectional} and TruthfulQA \citep{lin2022truthfulqa}), creativity (poem \citep{west2025base} and GuessBench \citep{zhu2025guessbench}), pluralism \citep{sorensen2024value}, sycophancy \citep{cheng2025social}, randomness, generation diversity (movie reviews \citep{wang2023detecting}), and uncertainty (AbstainQA \citep{feng2024don}), as supported by the findings of existing literature.
    \item \textcolor{Red}{Datasets where having the base model might be worse}: reasoning (gsm8k \citep{cobbe2021training}, BigBench-Hard \citep{suzgun2023challenging}, and NLGraph \citep{wang2023can}), instruction following (Alpaca \citep{dubois2023alpacafarm}), and safety (Coconot \citep{brahman2024art}), since these skills are explicitly what alignment is for.
    \item \textcolor{Dandelion}{Datasets where the effect of base models is unclear}: general QA (MMLU-pro \citep{wang2024mmlu}, AGIEval \citep{zhong2024agieval}, and PopQA \citep{mallen2023not}) and scientific literature (ScienceMeter \citep{wang2025sciencemeter}).
\end{itemize}
These datasets cover a wide range of LM capabilities that favor different model checkpoints in the training pipeline. If \ourmethod{} improves on \textcolor{Green}{category-1 tasks}, slightly behind/on par on \textcolor{Red}{category-2 tasks}, and on par/improves on \textcolor{Dandelion}{category-3 tasks}, it presents a promising collaboration strategy to fuse the strengths of model checkpoints and enable them to complement each other.

\section{Results}

We present the performance of individual models and model collaboration methods in Table \ref{tab:big}.

\begin{table}[t]
\vspace*{-30pt}
\caption{Performance of individual models and model collaboration methods. \textcolor{Green}{Green}, \textcolor{Red}{red}, and \textcolor{Dandelion}{yellow} denote category-\textcolor{Green}{1}/\textcolor{Red}{2}/\textcolor{Dandelion}{3} tasks in Section \ref{sec:experiment_settings}. $\downarrow$ denotes the lower the better. Best in \textbf{bold} and second-best in \underline{underline}. Model collaboration approaches outperform employing models individually on 16 out of 18 tasks. \ourmethod{} achieves the best performance on 13 tasks with a 12.9\% relative improvement over baselines on average.}
\label{tab:big}
\centering
\scriptsize
\setlength{\tabcolsep}{2pt}
\renewcommand{\arraystretch}{1.0}
\resizebox{1.0\linewidth}{!}{
\begin{tabular}{lccccccccc}\toprule[1.5pt]
&\textcolor{Green}{WikiDYK} &\textcolor{Green}{TruthfulQA} &\textcolor{Green}{Poem} &\textcolor{Green}{AbstainQA} &\textcolor{Green}{Pluralism} &\textcolor{Green}{Sycophancy} &\textcolor{Green}{GuessBench} &\textcolor{Green}{Numbers} ($\downarrow$) &\textcolor{Green}{Movie} \\\midrule[0.75pt]
\textsc{pretrained} &1.70 &10.37 &24.55 &\underline{62.92} &32.20 &12.80 &2.00 &\underline{0.26} &13.76 \\
\textsc{finetuned} &3.27 &30.63 &49.45 &49.44 &\bf 58.90 &13.00 &4.40 &0.90 &3.82 \\
\textsc{aligned} &3.92 &29.01 &\underline{77.70} &44.38 &50.90 &16.40 &6.40 &0.42 &0.93 \\ \midrule[0.75pt]
\textsc{prompt route} &3.27 &23.99 &54.40 &53.93 &45.80 &6.40 &7.60 &0.59 &4.94 \\
\textsc{routellm} &3.01 &\underline{34.38} &61.60 &58.43 &51.80 &14.20 &4.40 &0.87 &13.05 \\
\textsc{text collab} &3.14 &18.15 &52.30 &28.65 &28.30 &11.80 &6.80 &0.59 &3.24 \\
\textsc{text debate} &3.53 &16.86 &51.10 &30.90 &43.00 &6.40 &8.80 &0.54 &2.65 \\
\textsc{logit merge} &1.57 &9.08 &53.80 &12.92 &23.90 &14.80 &1.20 &\bf 0.22 &10.75 \\
\textsc{proxy tuning} &1.96 &2.76 &72.60 &26.40 &10.70 &15.20 &2.00 &0.37 &\bf 19.56 \\
\textsc{greedy soup} &\underline{4.58} &33.06 &\bf 77.95 &60.67 &\underline{57.80} &13.20 &6.80 &1.11 &13.91 \\
\textsc{dare ties} &2.61 &23.99 &47.90 &52.81 &37.50 &12.40 &1.60 &0.31 &10.60 \\ \midrule[0.75pt]
\textsc{switch-global} &4.44 &34.04 &70.90 &60.67 &55.60 &\underline{16.80} &\underline{10.80} &0.41 &14.14 \\
\textsc{switch-task} &\bf 5.75 &\bf 39.22 &70.25 &\bf 74.16 &53.20 &\bf 17.40 &\bf 13.60 &0.44 &\underline{17.39} \\ \midrule[1.5pt]
&\textcolor{Red}{GSM8k} &\textcolor{Red}{CocoNot} &\textcolor{Red}{Alpaca} &\textcolor{Red}{BBH} &\textcolor{Red}{NLGraph} &\textcolor{Dandelion}{MMLU-pro} &\textcolor{Dandelion}{AGIEval} &\textcolor{Dandelion}{PopQA} &\textcolor{Dandelion}{Science} \\ \midrule[0.75pt]
\textsc{pretrained} &27.20 &11.90 &14.51 &38.10 &44.50 &5.10 &4.41 &15.30 &29.80 \\
\textsc{finetuned} &37.20 &64.00 &52.43 &26.70 &38.33 &\underline{13.10} &11.94 &26.10 &56.30 \\
\textsc{aligned} &56.80 &53.10 &\bf 57.46 &35.20 &41.83 &10.40 &11.85 &31.20 &\underline{60.60} \\ \midrule[0.75pt]
\textsc{prompt route} &48.10 &43.60 &42.90 &43.40 &44.83 &5.10 &11.68 &29.90 &51.80 \\
\textsc{routellm} &48.10 &57.60 &43.97 &45.90 &48.67 &10.50 &12.32 &31.30 &59.00 \\
\textsc{text collab} &40.70 &41.80 &40.39 &34.60 &41.33 &7.10 &\underline{15.74} &31.30 &48.10 \\
\textsc{text debate} &46.90 &40.90 &52.35 &39.10 &44.67 &8.30 &15.05 &29.60 &52.70 \\
\textsc{logit merge} &38.30 &16.70 &28.46 &39.00 &25.67 &1.50 &7.01 &21.60 &10.50 \\
\textsc{proxy tuning} &44.50 &7.30 &37.27 &44.00 &43.83 &1.40 &2.34 &22.50 &11.40 \\
\textsc{greedy soup} &\underline{58.10} &66.00 &54.89 &36.50 &45.33 &12.70 &11.76 &31.30 &60.30 \\
\textsc{dare ties} &22.90 &19.40 &40.01 &30.30 &42.33 &6.50 &7.87 &25.10 &46.50 \\ \midrule[0.75pt]
\textsc{switch-global} &49.50 &\underline{70.90} &49.06 &\underline{52.60} &\underline{59.67} &13.00 &14.19 &\underline{33.70} &59.80 \\
\textsc{switch-task} &\bf 59.60 &\bf 72.80 &\underline{56.89} &\bf 58.30 &\bf 61.67 &\bf 16.70 &\bf 25.26 &\bf 37.70 &\bf 67.20 \\
\bottomrule[1.5pt]
\end{tabular}
}
\end{table}

\paragraph{Don't throw away your pretrained model.} Model collaboration among pretrained, finetuned, and aligned language models, baselines or ours, outperforms using these models individually on 16 of 18 tasks with 31.0\% relative improvement on average. This indicates that checkpoints other than the aligned models are diamonds in the rough, successfully complementing each other and contributing their unique strengths.

\begin{wraptable}{r}{0.6\textwidth}
\vspace*{-30pt}
\caption{Ablation study of generation patch size and switching strategy. Different tasks might favor different patch sizes while fine-tuning the switcher $f$ is consistently helpful.}
\vspace{10pt}
\centering
\scriptsize
\setlength{\tabcolsep}{2pt}
\renewcommand{\arraystretch}{1.0}
\label{tab:ablation_study}
\resizebox{1\linewidth}{!}{
\begin{tabular}{lccccc}\toprule[1.5pt]
Setting&TruthfulQA &Pluralism &GSM8k &BBH &PopQA \\\midrule[0.75pt]
\textsc{switch-task} &\bf 39.22 &53.20 &\bf 59.60 &\bf 58.30 &\bf 37.70 \\ \midrule[0.75pt]
\quad\textsc{patch size: 100} &30.31 &53.30 &44.70 &40.40 &32.00 \\
\quad\textsc{patch size: 30} &35.79 &55.70 &52.20 &53.50 &33.80 \\
\quad\textsc{patch size: 20} &32.58 &\bf 56.80 &47.80 &48.10 &31.80 \\
\quad\textsc{patch size: 10} &30.96 &54.50 &51.40 &48.70 &32.80 \\ \midrule[0.75pt]
\quad\textsc{random switch} &27.07 &54.80 &44.70 &53.10 &27.90 \\
\quad\textsc{untuned switch} &31.12 &53.10 &47.90 &41.80 &32.10 \\
\bottomrule[1.5pt]
\end{tabular}
}
\vspace*{-30pt}
\end{wraptable}

\paragraph{Switch Generation offers a strong collaboration strategy.} \ourmethod{} outperforms all individual models and model collaboration baselines on 13 datasets, with an average relative improvement of 12.9\%. In addition to improving on \textcolor{Green}{cat-1} tasks, \ourmethod{} also gains 6.58 points on average across \textcolor{Red}{cat-2} and \textcolor{Dandelion}{cat-3} tasks, where it was originally uncertain whether having the base model in collaboration might be helpful. This indicates that by collaborative inference with a flexible switching strategy, our approach adapts to diverse tasks through leveraging the strengths of candidate models and fusing their strengths.

\paragraph{Routing-based approaches are best for pretrained-aligned collaboration.} In descending order, routing-based, weight-based, text-based, and logit-based baselines achieve 31.15, 29.91, 26.32, and 18.97 points on average, indicating that routing-based methods are best suited for the collaboration of aligned and unaligned models, since different tasks require different skills that favor varying models. \ourmethod{} further provides a finer-grained and more flexible routing on the segment-level, so diverse model checkpoints could dynamically contribute in the problem-solving process when their skills are most needed.

\section{Analysis}
\label{sec:analysis}

\begin{wrapfigure}{r}{0.6\linewidth}
    \centering
    \vspace*{-10pt}
    \includegraphics[width=1\linewidth]{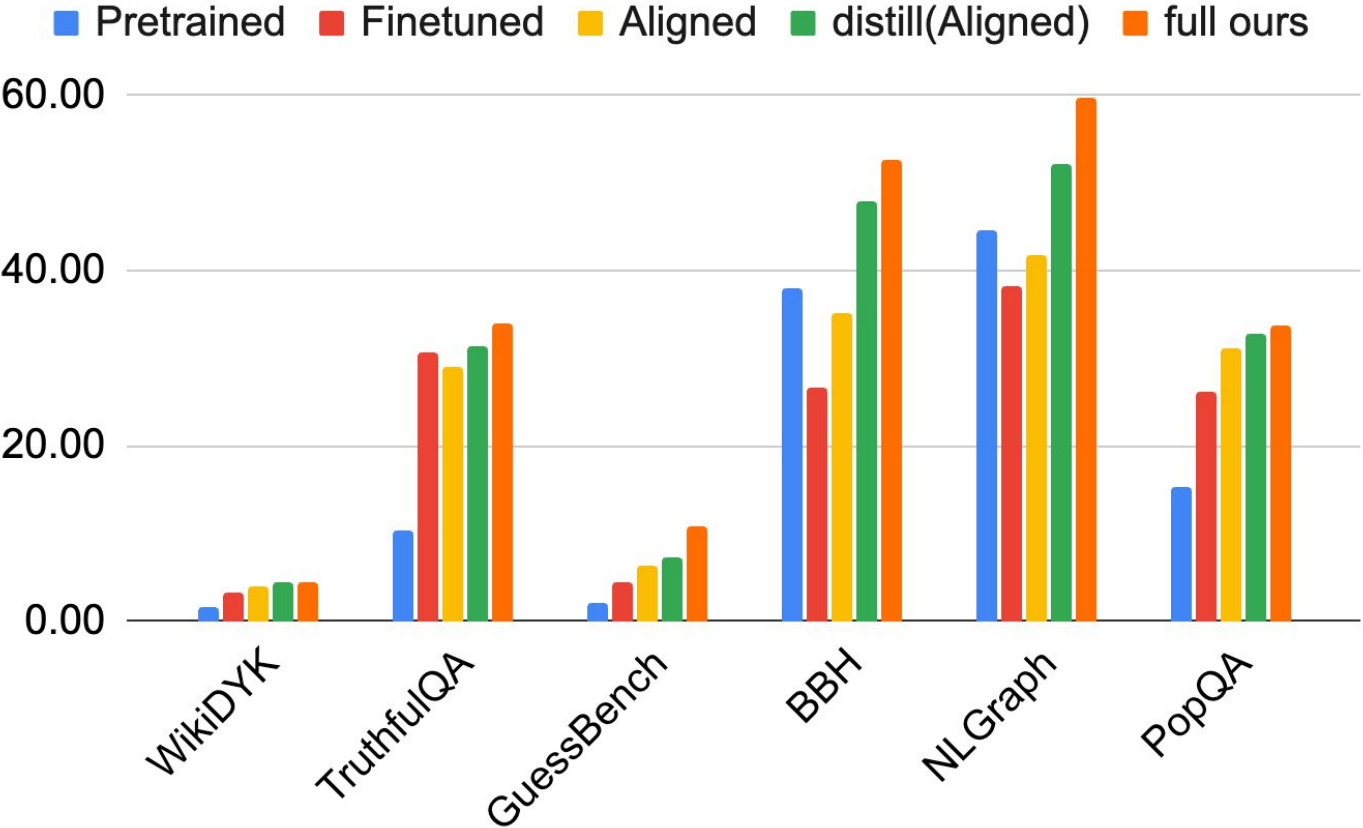}
    \caption{Distilling the collaboration patterns of \ourmethod{} back into the aligned model. Distillation recovers 58\% of the collaboration gains with only one fourth of the inference cost.}
    \label{fig:distill}
\end{wrapfigure}

\paragraph{Ablation study} We conduct ablation study on two key design choices in \ourmethod{}: 1) generation patch size: we by default generate 50 tokens per model and call the switcher $f$ and we additionally employ \{10, 20, 30, 100\} in this study; 2) switcher training: we by default train the switcher $f$ through supervised fine-tuning on simulated switching outcomes and we additionally employ random switching ($f_\textit{random} = \mathrm{Uniform}(n)$) and untuned switching (directly employing the aligned model as $f$ without fine-tuning). Results in Table \ref{tab:ablation_study} demonstrate that different tasks might benefit from different switching granularity: by employing more frequent switching on Pluralism our approach further improves. Fine-tuning the switcher with our methodology is consistently effective as it outperforms random and untuned switching on all five tasks.

\begin{wrapfigure}{r}{0.5\linewidth}
    \centering
    \vspace*{-10pt}
    \includegraphics[width=1\linewidth]{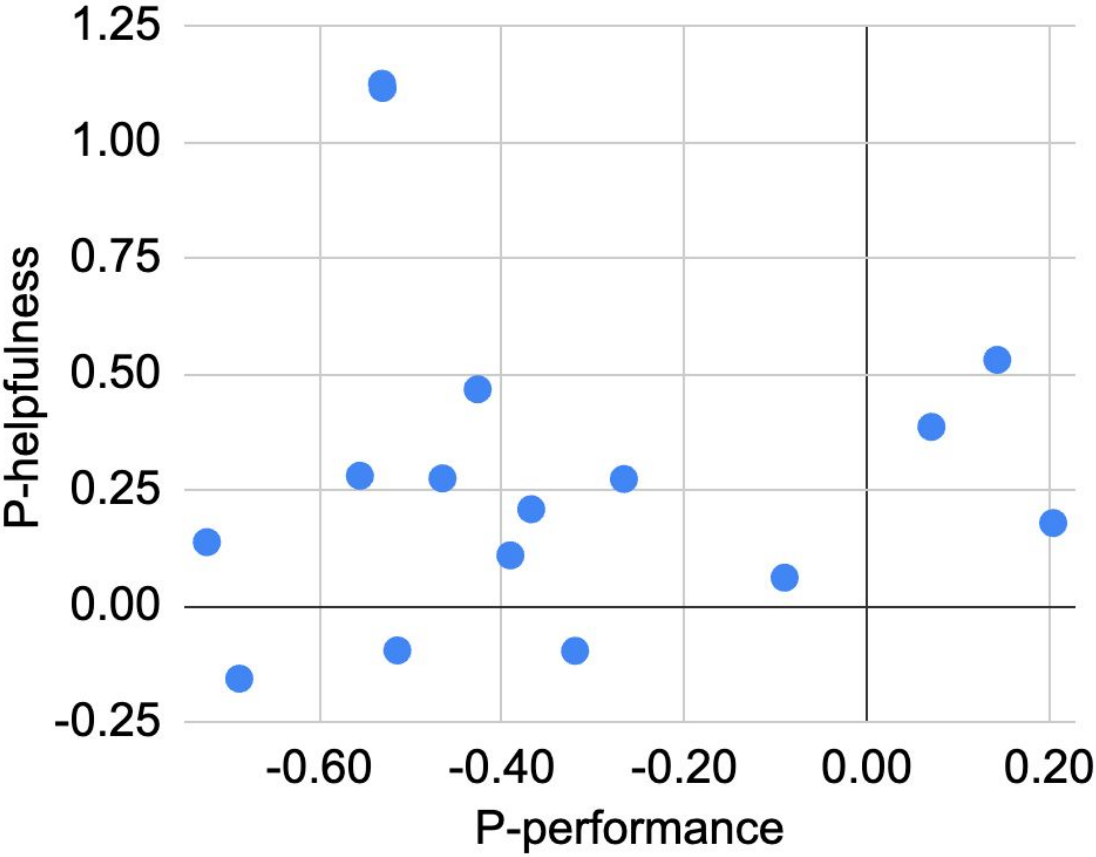}
    \vspace*{-10pt}
    \caption{Correlation between the performance and helpfulness of the pretrained model. While not being the best individual model, it is consistently helpful in the model collaboration system.}
    \label{fig:performance_helpfulness}
\end{wrapfigure}

\paragraph{Distillation back into a single model} At inference time, \ourmethod{} loads and generates texts with n+1 LMs simultaneously. While this is fast with multiple GPUs and multiprocessing, we propose to reduce inference costs by \emph{distilling the switching patterns back into a single model}. The aligned model was once pretrained and finetuned too, so by distillation we hope that it could recover the submerged capabilities of strengths of its previous forms. For inputs in a dataset, we 1) generate outputs with full \ourmethod{}, 2) fine-tune the aligned model on the generated outputs, and 3) evaluate the performance of the distilled aligned model when used individually. Results in Figure \ref{fig:distill} demonstrate that distillation successfully helps the aligned model pick up the submerged skills, recovering 57.5\% of the gains of \ourmethod{} with only one fourth of the inference cost (one model only vs. three models and a switcher LM). This success sheds light on the broader potential of distilling multi-model/agent systems back into a single model/agent for inference-time efficiency.

\paragraph{Correlation between individual performance and helpfulness} While the pretrained base model has many strengths, it consistently isn't the best-performing individual model in Table \ref{tab:big}. However, this doesn't prevent it from being helpful in the collaboration and contributing its strengths when needed. We quantify this phenomenon with two metrics for each task: $\mathrm{P\mhyphen performance} = \frac{P-\max(P,F,A)}{\max(P,F,A)}, \ \ \ \mathrm{P\mhyphen helpfulness} = \frac{C(P,F,A)-\max(P,F,A)}{\max(P,F,A)}$, where $P, F, A$ indicate the performance of pretrained, finetuned, aligned models when employed individually and $C(P,F,A)$ indicates the performance of their collaboration (through \ourmethod{}). Results in Figure \ref{fig:performance_helpfulness} demonstrate that the vast majority of tasks fall into the top-left quadrant: while the pretrained model isn't the best when employed individually ($\mathrm{P\mhyphen performance} < 0$), \ourmethod{} leverages its strengths to gain in collaboration ($\mathrm{P\mhyphen helpfulness} > 0$). This highlights the broader potential that weak models are not useless: they are rightfully diamonds in the rough and contribute their unique strengths when employed in the right model collaboration system.

\begin{table}[t]
\caption{Performance when directly employing the trained switcher model for other model settings: PFA from another model family, PA only, PF and two versions of A, and three specialized LM experts. The trained switcher model consistently generalizes to these settings to varying extents.}
\vspace{10pt}
\label{tab:generalize_model}
\centering
\scriptsize
\setlength{\tabcolsep}{2pt}
\renewcommand{\arraystretch}{1.0}
\resizebox{1.0\linewidth}{!}{
\begin{tabular}{lccc|ccc|ccc|ccc}\toprule[1.5pt]
&\multicolumn{3}{c}{Setting 1} &\multicolumn{3}{c}{Setting 2} &\multicolumn{3}{c}{Setting 3} &\multicolumn{3}{c}{Setting 4} \\\cmidrule{2-13}
&Truthful &BBH &PopQA &Truthful &BBH &PopQA &Truthful &BBH &PopQA &Truthful &BBH &PopQA \\\midrule
\textsc{model \#1} &26.09 &69.50 &24.10 &10.37 &38.10 &15.30 &10.37 &38.10 &15.30 &40.19 &48.40 &20.90 \\
\textsc{model \#2} &53.00 &54.80 &25.60 &29.01 &35.20 &31.20 &30.63 &26.70 &26.10 &53.00 &54.80 &25.60 \\
\textsc{model \#3} &63.29 &70.70 &23.30 &/ &/ &/ &32.58 &36.20 &31.10 &54.29 &56.20 &22.20 \\
\textsc{model \#4} &/ &/ &/ &/ &/ &/ &29.01 &35.20 &31.20 &/ &/ &/ \\
\textsc{ours} &\textbf{66.64} &\textbf{73.70} &\textbf{27.60} &\textbf{32.58} &\textbf{45.60} &\textbf{34.60} &\textbf{35.82} &\textbf{48.70} &\textbf{31.70} &\textbf{55.27} &\textbf{57.60} &\textbf{26.90} \\
\bottomrule[1.5pt]
\end{tabular}
}
\end{table}

\paragraph{Generalizing to unseen models} The trained switcher LM $f$ (in switch-global) has learned from diverse tasks, contexts, and model collaboration patterns among the Tulu-v3 suite of models. We hypothesize that $f$ could be employed off-the-shelf for switch generation with other model settings, in increasing difficulty and generalization gap:

\begin{itemize}[leftmargin=*]
    \item \emph{Setting 1}: pretrained, finetuned, and aligned models in another model family, specifically Qwen2.5-7B \citep{qwen2}.
    \item \emph{Setting 2}: one fewer model: only pretrained and aligned of Tulu-v3.
    \item \emph{Setting 3}: one more model: pretrained, finetuned, and two versions of aligned (DPO and RLVR) of Tulu-v3.
    \item \emph{Setting 4}: three specialized LM experts \citep{jiang2025sparta} that are not the aligned version of each other.
\end{itemize}

Table \ref{tab:generalize_model} shows that the switcher consistently generalizes to these four settings, with an average relative improvement of 5.8\%, 14.3\%, 13.1\%, and 3.1\%. We will release the switcher model $f$ as an artifact and encourage readers to employ it for switch generation with their suite of models.

\begin{wraptable}{r}{0.6\textwidth}
\vspace*{-20pt}
\caption{The trained swicher model generalizes to unseen tasks, outperforming baselines on most tasks.}
\vspace{2pt}
\centering
\scriptsize
\setlength{\tabcolsep}{2pt}
\renewcommand{\arraystretch}{1.0}
\label{tab:generalize_task}
\resizebox{1\linewidth}{!}{
\begin{tabular}{lcccccc}\toprule[1.5pt]
&Normad &Interests &MATH &K-Cross &ARC &MedQA \\\midrule[0.75pt]
\textsc{pretrained} &29.85 &43.47 &27.89 &7.00 &16.13 &20.30 \\
\textsc{finetuned} &46.40 &63.30 &25.75 &24.60 &45.22 &26.50 \\
\textsc{aligned} &48.70 &66.95 &28.74 &20.80 &46.76 &28.10 \\
\textsc{text debate} & 34.90 & \bf 70.10 & \underline{34.55} & 12.00 & 30.97 & 28.50 \\
\textsc{greedy soup} & \underline{48.85} & 67.35 & 29.41 & \underline{25.50} & \underline{47.44} & \underline{28.90} \\
\textsc{switch-global} &\bf 50.65 &\underline{69.78} &\bf 37.36 &\bf  25.70 &\bf  48.38 &\bf  31.50 \\
\bottomrule[1.5pt]
\end{tabular}
}
\vspace*{-20pt}
\end{wraptable}

\paragraph{Generalizing to unseen tasks} We directly employ the trained switcher LM $f$ (in switch-global) and compare it against two strong baselines on six additional tasks spanning the three task categories (Normad \citep{rao-etal-2025-normad}, human interests \citep{feng2025model}, MATH \citep{hendrycks2020measuring}, Knowledge Crosswords \citep{ding-etal-2024-knowledge}, ARC-challenge \citep{clark2018think}, and MedQA \citep{jin2021disease}). Results in Table \ref{tab:generalize_task} demonstrate that by learning from the switching patterns across diverse tasks, \ourmethod{} generalizes to unseen tasks and outperforms baselines by 3.9\% on average.

\begin{figure}[t]
    \centering
    \vspace*{10pt}
    \includegraphics[width=1\linewidth]{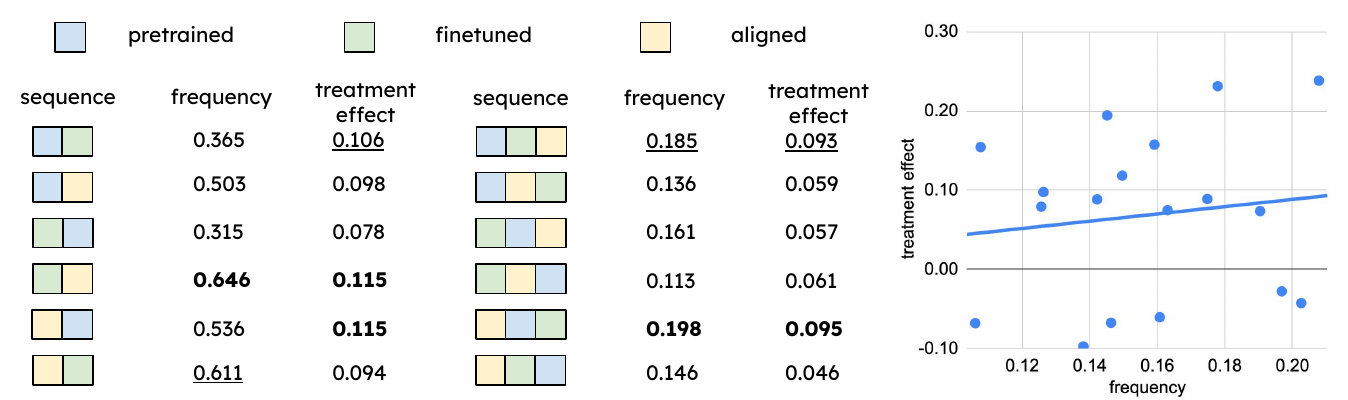}
    \caption{Frequency and treatment effect of 2-length (left) and 3-length (middle) switching sequences, as well as their correlation across three task categories for 3-length sequences (right). \ourmethod{} learns to identify helpful switching patterns and frequently leverages them.}
    \label{fig:good_sequence}
\end{figure}

\begin{wraptable}{r}{0.6\textwidth}
\vspace*{-20pt}
\caption{\ourmethod{} solves 10.7\% problems that no individual model did. Color shades denote outcomes that \textcolor{ForestGreen}{discover new skills}, \textcolor{LimeGreen}{retain existing skills}, \textcolor{Dandelion}{no change}, and might \textcolor{Red}{lose skills}.}
\vspace{2pt}
\centering
\scriptsize
\setlength{\tabcolsep}{2pt}
\renewcommand{\arraystretch}{1.0}
\label{tab:union_of_correctness}
\resizebox{1\linewidth}{!}{
\begin{tabular}{lcccccc}\toprule[1.5pt]
&\multicolumn{2}{c}{Truthful} &\multicolumn{2}{c}{AbstainQA} &\multicolumn{2}{c}{BBH} \\\cmidrule{2-7}
PFA / ours &correct &wrong &correct &wrong &correct &wrong \\\midrule[0.75pt]
all correct &\cellcolor[HTML]{d9ead3}3.6\% &\cellcolor[HTML]{f4cccc}0.3\% &\cellcolor[HTML]{d9ead3}15.7\% &\cellcolor[HTML]{f4cccc}2.2\% &\cellcolor[HTML]{d9ead3}7.4\% &\cellcolor[HTML]{f4cccc}0.9\% \\
$\geq$1 correct &\cellcolor[HTML]{d9ead3}25.3\% &\cellcolor[HTML]{f4cccc}7.6\% &\cellcolor[HTML]{d9ead3}50.6\% &\cellcolor[HTML]{f4cccc}5.1\% &\cellcolor[HTML]{d9ead3}37.1\% &\cellcolor[HTML]{f4cccc}8.6\% \\
all wrong &\cellcolor[HTML]{93c47d}10.4\% &\cellcolor[HTML]{fff2cc}52.8\% &\cellcolor[HTML]{93c47d}7.9\% &\cellcolor[HTML]{fff2cc}18.5\% &\cellcolor[HTML]{93c47d}13.8\% &\cellcolor[HTML]{fff2cc}32.2\% \\
\bottomrule[1.5pt]
\end{tabular}
}
\vspace{-10pt}
\end{wraptable}

\paragraph{Good sequences} By running \ourmethod{} across 18 tasks, we accumulate valuable traces of model collaboration and switching patterns. Within them exist many \emph{switching sequences} (e.g. \emph{``PFA''}: pretrained generates first, followed by finetuned, followed by aligned): if we could identify which of these sequences are \emph{good}, we could 1) directly employ these switching patterns off-the-shelf without calling the switcher LM for efficiency and/or 2) steer \ourmethod{} towards employing these sequences more often. We define two metrics for switching sequences:

\begin{itemize}[leftmargin=*]
    \item \emph{Frequency}: in what percentage of responses was this sequence employed?
    \item \emph{Treatment effect}: performance when this sequence is employed minus when not employed.
\end{itemize}

We present results for all unique 2-length and 3-length sequences as well as their correlation in Figure \ref{fig:good_sequence}: the most helpful sequences (with the highest treatment effect) are also among the most frequent, and there is a positive correlation between the two with $R^2=0.017$. This indicates that \ourmethod{} learns to identify helpful switching patterns and more frequently leverage them for better collaboration.

\begin{wrapfigure}{r}{0.5\linewidth}
    \centering
    \vspace*{-30pt}
    \includegraphics[width=1\linewidth]{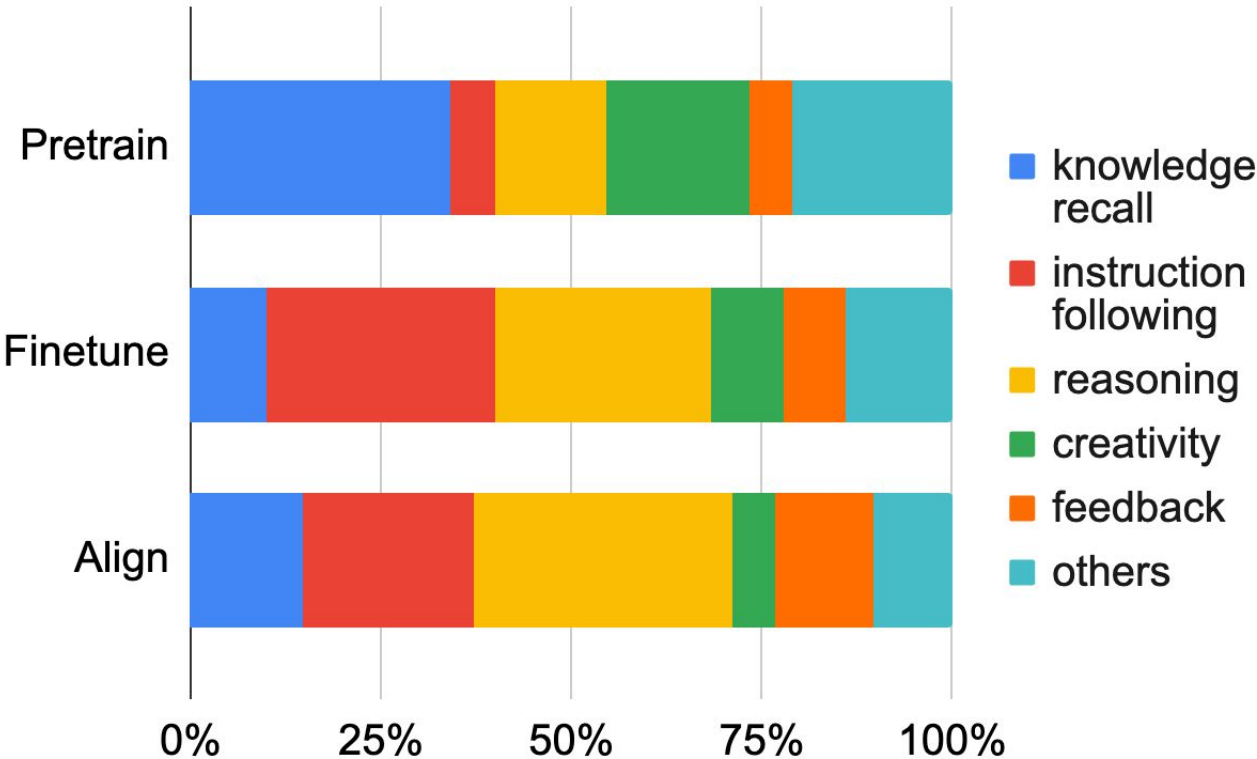}
    \caption{Roles that each model plays in their generated segments, averaged across all datasets.}
    \label{fig:model_roles}
    \vspace*{-20pt}
\end{wrapfigure}

\paragraph{New skills} We hypothesize that the performance gains of \ourmethod{} might come from two aspects: 1) aggregate skills that one of the models already has, and 2) solving problems that none of the models could solve individually. We present the statistics between single-model and multi-model correctness in Table \ref{tab:union_of_correctness}: \ourmethod{} successfully answers 10.7\% of problems that none of the individual models could, while only losing out on 8.2\% of problems that one model could individually solve, netting a benefit of 2.5\% through model collaboration.

\paragraph{Model Roles} Model responses often feature a wide range of skills that favor different model stages (Figure \ref{fig:teaser}): we investigate under \ourmethod{}, what are the roles of the pretrained, finetuned, and aligned langauge models in generated texts? Specifically, we first identify a suite of five skills with high frequency (knowledge recall, instruction following, reasoning, creativity, and feedback). We then employ LLM-as-a-judge \citep{zheng2023judging}, specifically the GPT-4o model, to annotate each generated segment for one of the skills (or ``others'') given the query, trace, and the full generated sequence. We manually examine 100 LLM annotations and find an 86\% agreement between human-LLM judgements (with the most difference in the ``others'' category). We average it across datasets and report the results in Figure \ref{fig:model_roles}: it shows that models are largely performing the skills that they are good at: for example, the pretrained model is most frequently used for knowledge recall, while the aligned model is most used for reasoning. This indicates that \ourmethod{} and the trained switcher $f$ learns to leverage model strengths when their skills are most needed.

\vspace*{-10pt}
\section{Related Work}
\vspace*{-10pt}

\paragraph{The Tradeoffs of Alignment} Alignment and reinforcement learning have become an indispensable part of language model training: they are credited for valuable skills in state-of-the-art LMs such as reasoning \citep{guo2025deepseek}, safety \citep{zhang2024r}, agentic applications \citep{ma2024coevolving}, and more \citep{ouyang2022training}. An increasing line of research recognizes that \emph{alignment has tradeoffs} \citep{lin2024mitigating}, that the pretrained and unaligned base models might have advantages on skills such as creativity \citep{west2025base}, uncertainty \citep{tian2023just}, pluralism \citep{feng2024modular}, knowledge \citep{wang2025sciencemeter}, or even reasoning itself \citep{yue2025does}. However, we couldn't directly employ the base model for these domains since they struggle to follow instructions and lack safety guardrails. We propose to make the best of both worlds by \emph{not throwing away your base model} and instead leveraging model collaboration across diverse checkpoints in the training pipeline to fuse model strengths and complement each other.

\paragraph{Model Collaboration} Advancing beyond training a single, generalist language model, recent research is increasingly emphasizing modularity through \emph{model collaboration}, where diverse (language) models collaborate, compose, and complement each other \citep{feng2025one}. Model collaboration approaches mainly vary by the level of information exchange: API-level methods such as routing \citep{ongroutellm, frick2025prompt, fenggraphrouter, zheng2025citer} and cascading \citep{chenfrugalgpt, guptalanguage, yuelarge}, text-level methods through collaboration \citep{feng2024don, guo2024large, zhao2025language, dang2025multi} or competition \citep{du2023improving, liang2024encouraging, zhaocompeteai}, logit-level methods with logit fusion or contrast \citep{pei2023preadd, li2023contrastive, mavromatispack, chuangdola, mitchellemulator, liutuning, huangdivide}, and weight-level methods such as model merging \citep{yadav2023ties, yu2024language, huanglorahub, feng2025model, zengprobabilistic} and Mixture-of-Experts \citep{sukhbaatarbranch, diao2023mixture, yadavsurvey, shi2025flexolmo}. Since model responses are often not monolithic, featuring a diverse set of skills that favor different model stages (Figure \ref{fig:teaser}), we propose \ourmethod{} for the collaborative inference of pretrained, finetuned, and aligned LMs where they take turns to generate in a response sequence. \ourmethod{} is related to various model collaboration protocols \citep{fei2024nudging, shen2024learning, wangspeculate} while uniquely training a switcher \emph{LM} as the switching strategy, switching by the granularity of patches, and offers generalization to unseen models as switching candidates. Our work also highlights that we don't need to always train new models for collaboration: byproducts in existing model development lifecycles could be reused and repurposed for new potential.

\vspace*{-10pt}
\section{Conclusion}
\vspace*{-10pt}
We propose \ourmethod{}, an inference-time model collaboration strategy where multiple models in the training pipeline are dynamically selected to generate text in a single response. By training and employing a switcher LM, multiple models dynamically generate text segments and contribute their strengths when most needed. Extensive experiments demonstrate that \ourmethod{} outperforms each individual constituent models and eight model collaboration baselines on 13 datasets by 12.9\% on average. Further analysis reveals that \ourmethod{} generalizes to unseen models and tasks, as well as identifying and frequently employing helpful collaboration patterns. Our work uniquely highlights the huge potential of reusing by-product models and checkpoints in current LM training pipelines that are otherwise discarded.

\bibliography{iclr2026_conference}
\bibliographystyle{iclr2026_conference}

\appendix

\section*{Limitations}

\ourmethod{} jointly employs multiple model checkpoints in the training pipeline for collaborative inference, which incurs extra cost (compared to just using the aligned version) in exchange for compositional model strengths. The extra cost could be mitigated on several fronts: 1) by employing multiple GPUs and multiprocessing for parallel text generation over batches of instructions, the throughput is much higher compared to using a single model; 2) by calling the switching strategy every patch (instead of every token), the switching overhead is significantly reduced and the user could also configure the patch size to customize the cost; 3) by distilling the collaboration patterns in \ourmethod{} back into the aligned model (Figure \ref{fig:distill}), we recover part of the performance gains while cutting inference costs back to a single model.

We observe that switching by patches works better than tokens, so we employ fixed-size patches in \ourmethod{}. We also observe that different tasks might need different amounts of generated tokens; thus, the optimal patch size might also change across tasks and contexts. We treat it as a hyperparameter for now: future work could look into flexible and dynamic adjustments of patch sizes and switching frequency.

\section*{Reproducibility Statement} We provide extensive experiment details such as hyperparameter settings, dataset statistics, and more in Section \ref{sec:experiment_settings} and Appendix \ref{sec:experiment_details}. We will release the training and inference code, switcher LMs, and experiment logs upon acceptance.

\section*{Ethics Statement}

\ourmethod{} is a model collaboration protocol across multiple language models, so it is susceptible to malicious contributions: for example, if the alignment datastore is compromised and the aligned model is malicious, when used in collaboration, the system would also be seriously impacted. Safety in model collaboration systems is a critical future research question, and its findings would have great impacts on \ourmethod{}.

\begin{wrapfigure}{r}{0.5\linewidth}
    \centering
    \vspace*{-20pt}
    \includegraphics[width=1\linewidth]{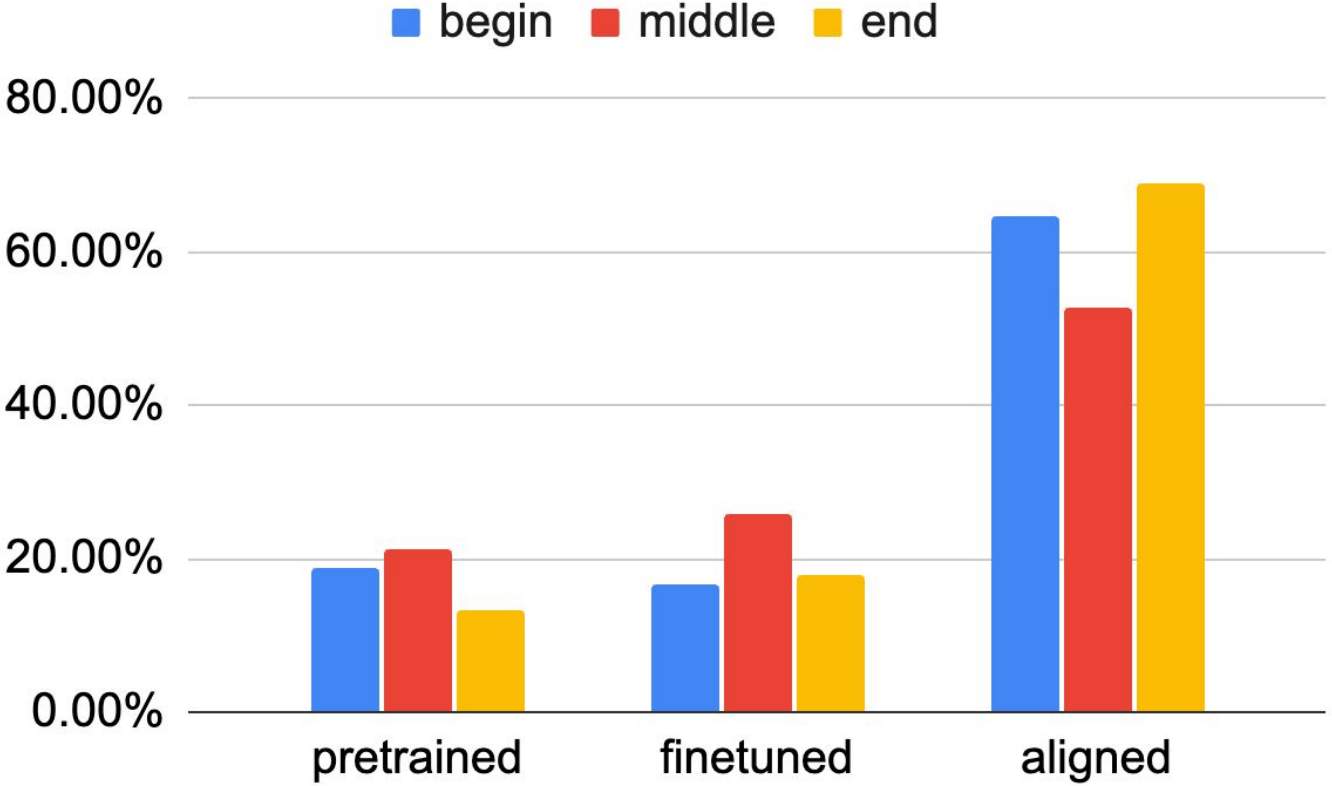}
    \caption{Frequency of pretrained, finetuned, and aligned models being used in the begin/middle/end of the sequence.}
    \vspace*{-40pt}
    \label{fig:locations}
\end{wrapfigure}

As the base (and finetuned) versions of models are mostly not safety-aligned, it is possible that having them in collaboration might override the safety guardrails of aligned language models. This might have implications for red teaming and adversarial language modeling.

\section{Analysis (cont.)}

\paragraph{Model Locations} We investigate whether pretrained, finetuned, and aligned language models might be used more frequently in the beginning (first one-third), middle (one-third to two-thirds), and the end (two-thirds to end) patches in \ourmethod{}. Results in Figure \ref{fig:locations} demonstrate that the aligned model is more frequently employed in the beginning and the end while the pretrained/finetuned model is more employed in the middle, suggesting that the middle of the response is more suited for exploration, while the beginning/end requires instruction following and summarization that favors the aligned model.

\begin{wrapfigure}{r}{0.5\linewidth}
    \centering
    \includegraphics[width=1\linewidth]{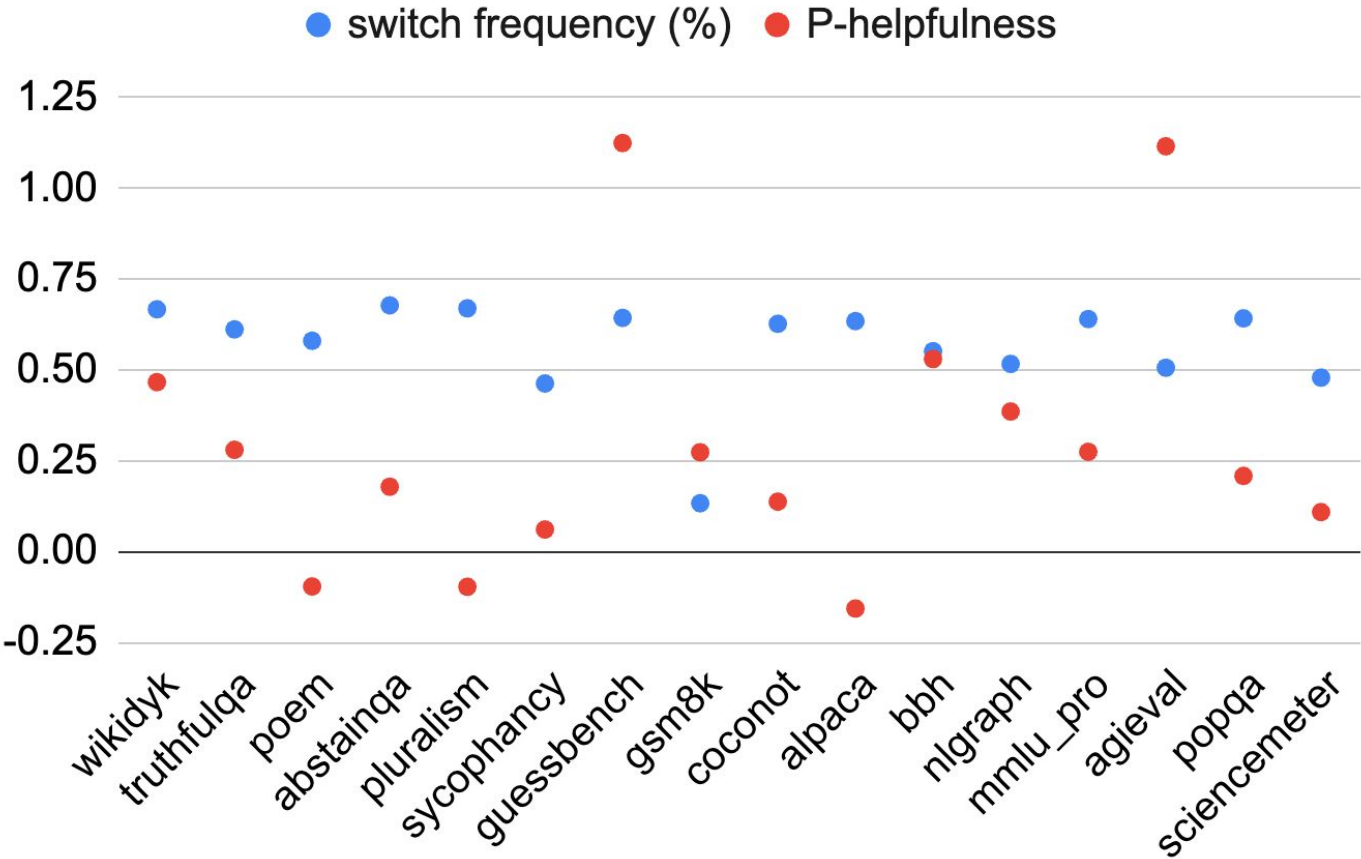}
    \caption{Switching frequency and P-helpfulness across tasks.}
    \vspace*{-10pt}
    \label{fig:switching_frequency}
\end{wrapfigure}

\paragraph{Switching Frequency} How often does the switcher $f$ decide that it's time to change to another model? We plot the switching frequency and collaboration helpfulness (P-helpfulness, Figure \ref{fig:performance_helpfulness}) in Figure \ref{fig:switching_frequency}: it is demonstrated that the switching frequency is consistently high, indicating that the models are actively used in collaboration. There isn't a consistent conclusion about whether more or less switching is better for performance.

\paragraph{Qualitative Analysis} We present examples where the pretrained, finetuned, and aligned models did not generate a good response individually, while \ourmethod{} was successful in generating a good response through collaboration in Tables \ref{tab:qualitative_1} to \ref{tab:qualitative_3}. It shows that \ourmethod{} has better deliberation and more extensive ``reasoning'' and explanation to reach a more well-rounded response. This also suggests that while the pretrained and finetuned models might not be following instructions when used individually, if the aligned model provides good context to work on, they will be helpful.
\section{Experiment Details}
\label{sec:experiment_details}

\begin{wraptable}{r}{0.5\textwidth}
\centering
\scriptsize
\setlength{\tabcolsep}{3pt}
\renewcommand{\arraystretch}{1}
\vspace*{-10pt}
\resizebox{1\linewidth}{!}{
\begin{tabular}{lccc}
\toprule[1.5pt]
\multirow{2}{*}{Dataset} &\multirow{2}{*}{Source} &\multicolumn{2}{c}{Size} \\\cmidrule{3-4}
& &dev &test \\\midrule
Sycophancy & \citep{cheng2025social} & 1000 & 1000 \\
AbstainQA** & \citep{feng2024don} & 178 & 178 \\
Normad & \citep{rao-etal-2025-normad} & 500 & 2000 \\
ScienceMeter*** & \citep{wang2025sciencemeter} & 1000 & 1000 \\
MATH* & \citep{hendrycks2020measuring} & 956 & 956 \\
Human Interests & \citep{feng2025model} & 400 & 400 \\
AGIEval*** & \citep{zhong2024agieval} & 1156 & 1156 \\
CocoNot*** & \citep{brahman2024art} & 1000 & 1000 \\
TruthfulQA** & \citep{lin2022truthfulqa} & 200 & 617 \\
WikiDYK & \citep{zhang2025bidirectional} & 6849 & 765 \\
MMLU-pro** & \citep{wang2024mmlu} & 70 & 1000 \\
BBH*** & \citep{suzgun2023challenging} & 1000 & 1000 \\
PopQA*** & \citep{mallen2023not} & 1000 & 1000 \\
K-Crosswords & \citep{ding-etal-2024-knowledge} & 200 & 1000 \\
GuessBench** & \citep{zhu2025guessbench} & 250 & 250 \\
Movies & \citep{wang2023detecting} & 200 & 200 \\
GSM8k & \citep{cobbe2021training} & 200 & 1000 \\
Alpaca & \citep{dubois2023alpacafarm} & 10000 & 1000 \\
NLGraph*** & \citep{wang2023can} & 600 & 600 \\
\bottomrule[1.5pt]
\end{tabular}
}
\caption{Statistics of employed datasets. *, **, and *** indicate the improvement of \ourmethod{} against baselines for this dataset is statistically significant with $p<0.1$, $p<0.05$, and $p<0.01$ with one-tailed z-test.}
\vspace*{-50pt}
\label{tab:dataset_statistics}
\end{wraptable}

\paragraph{Dataset Details} We employ 18+6 diverse datasets spanning multiple LM capability areas for evaluation in this work. All datasets are evaluated with zero-shot prompting. Sycophancy uses the original prompt in \citet{cheng2025social} and GPT-4o for evaluation: only if a response does not fall into any of the sycophantic categories we give a score of 1, otherwise 0. AbstainQA employs abstain accuracy \citep{feng2024don} as the evaluation metric. ScienceMeter is employed as an NLI task where the passage either supports or refutes the claim. CocoNot \citep{brahman2024art} uses the regex in the original paper to judge contextual non-compliance. We employ the multiple-choice setting of TruthfulQA. We employ GPT-4o to generate a description of images in GuessBench \citep{zhu2025guessbench} to transform it into a language task. Movie reviews are generated ten times for each movie, five times with the IMDB summary and five times without, employing RoBERTa-base \citep{liu2019roberta} for embeddings, and calculating the average pairwise distance. NLGraph \citep{wang2023can} are evaluated with 50\% connectivity and 50\% shortest path problems. Statistics and statistical significance tests are presented in Table \ref{tab:dataset_statistics}.

\paragraph{Hyperparameter Details} We descibe main hyperparameter configurations in Section \ref{sec:experiment_settings}. We run grid search for training the switcher with  $\text{epoch} \in \{1,2,3,4,5,6,7,8\}$ and learning rate $l \in \{1e-3, 5e-4, 2e-4, 1e-4, 5e-5\}$. We then select the switcher $f$ that leads to the best performance on the dev set for evaluation on the test set.

\paragraph{Baseline Details} Pretrained, finetuned, and aligned models used individually with $top-p=0.7$ and 512 maximum generated tokens (same as \ourmethod{}); \textsc{prompt route} employs the aligned model and the following prompt to select one of the three models for inference: \textit{``You are given an instruction and three candidate models: a pretrained model, a supervised fine-tuned model, and an aligned model. Please select a model that might best fulfill the instruction. Instruction: [instruction] You are given an instruction and three candidate models: a pretrained model, a supervised fine-tuned model, and an aligned model. Please select a model that might best fulfill the instruction. Response:''}; \textsc{routellm} employs the ``causal LLM classifier'' approach in \citet{ongroutellm} with the aligned model; \textsc{text collaborate} employs the prompts in \citet{feng2024don}; \textsc{text debate} employs the prompts in \citet{du2023improving}; \textsc{logit merge} and \textsc{proxy tuning} both employs the implementation of \citet{liutuning}, with the former approach assigns a weight of 1/3 to every model and the second is (best + (best - worst)); \textsc{greedy soup} and \textsc{dare-ties} employ the implementation at MergeKit \citep{goddard-etal-2024-arcees}.

\newpage

\begin{table}[t]
    \centering
    \small
    \setlength{\tabcolsep}{2pt}
    \renewcommand{\arraystretch}{1.1}
    \resizebox{1.0\linewidth}{!}{
    \begin{tabularx}{\linewidth}{m|X}
        \toprule[1.5pt]
        Instruction & What location in the West Bank has been associated with the temptation of Christ since at least the 4th century? \\ \midrule[0.75pt]
        Pretrained &  The Church of the Holy Sepulchre. \xmark \\ \midrule[0.75pt]
        Finetuned & Jericho \xmark \\ \midrule[0.75pt]
        Aligned & The location in the West Bank associated with the temptation of Christ is traditionally identified as the "Place of the Sermon" or "Petrine Monastery" on the top of the Mount of Olives. It is believed that this is where Jesus was tempted by the Devil, according to the Gospels of Matthew, Mark, and Luke. The tradition holds that the actual place where the temptation took place was on the nearby rocky terrain known as the Garden of Gethsemane, located just below the Mount of Olives. The Mount of Olives is a rocky hill east of Jerusalem, and the Garden of Gethsemane is where Jesus is traditionally believed to have prayed before his arrest. \xmark \\ \midrule[0.75pt]
        Switch Generation & The location in the West Bank associated with the temptation of Christ since at least the 4th century is traditionally considered to be the Mount of Temptation, also known as the Quarantine. According to Christian tradition, it is believed that Jesus was 40 days in the desert, tempted by the devil, and fasted. This mountain is located in the Judaean Desert, between the Dead Sea and Jerusalem. The mountain is approximately 2,700 feet above sea level, and is a popular pilgrimage site for Christians. The site is marked by a chapel, which dates back to the 4th century and was built to commemorate the event. The chapel has undergone various renovations and modifications over the centuries, but the basic structure remains a symbol of this significant event in Christian belief. It's important to note that the Mount of Temptation is within the West Bank, an area that has been a point of contention in the ongoing Israeli-Palestinian conflict. Access to and control over religious sites such  as the Mount of Temptation have been a source of tension and conflict between the two groups. \cmark \\
        \bottomrule[1.5pt]
    \end{tabularx}
    }
    \caption{Example one of individual model and \ourmethod{} responses.}
    \label{tab:qualitative_1}
\end{table}

\begin{table}[t]
    \scriptsize
    \small
    \setlength{\tabcolsep}{2pt}
    \renewcommand{\arraystretch}{1.1}
    \resizebox{1.0\linewidth}{!}{
    \begin{tabularx}{\linewidth}{m|X}
        \toprule[1.5pt]
        Instruction & Passage 1.At the 2007 American Geophysical Union’s.meeting in Acapulco, Mexico, some two dozen.scientists presented multiple studies arguing that a.comet or asteroid exploded above or on the northern.ice cap almost 13,000 years ago—showering debris.across the North American continent and causing.temperatures to plunge for the next millennium..The team argues that its idea explains multiple.observations: not only the climate cooling and the.disappearance of the Clovis hunters, but also the.near-simultaneous extinction of the continent’s large.mammals..Not all will be convinced. Several leading.hypotheses already explain each of these three events..A change in ocean circulation is generally thought to.have brought about the onset of the millennium-long.cooling, which is known as the Younger Dryas. This.cooling might, in turn, have caused the Clovis.hunters to disappear. And, if they had not previously.been killed by disease or hunted to extinction, the big.prehistoric beasts may also have been doomed by this.change in climate..The new evidence comes in the form of.geochemical analysis of sedimentary layers at 25.archaeological sites across North America—9 of.them Clovis. Certain features of the layers, say the.team, suggest that they contain debris formed by an.extraterrestrial impact. These include spherules of.glass and carbon, and amounts of the element.iridium said to be too high to have originated on.Earth. In addition, the rocks contain black layers of.carbonized material, which the team says are the.remains of wildfires that swept across the continent.after the impact..Passage 2. (...) The team argues that when the quality.and accuracy of the dating—which was based on.radiocarbon and other techniques—is examined.closely, only three of the 29 sites actually fall within.the time frame of the Younger Dryas onset, about 12,800 years ago; the rest were probably either earlier.or later by hundreds (and in one case, thousands) of.years..“The supposed Younger Dryas impact fails on.both theoretical and empirical grounds,” says.Meltzer, who adds that the popular appeal of the.hypothesis is probably due to the way that it provides.“simple explanations for complex problems.” Thus,.“giant chunks of space debris clobbering the planet.and wiping out life on Earth has undeniably broad.appeal,” Meltzer says, whereas “no one in Hollywood.makes movies” about more nuanced explanations,.such as Clovis points disappearing because early.Americans turned to other forms of stone tool.technology as the large mammals they were hunting.went extinct as a result of the changing climate or.hunting pressure..But impact proponents appear unmoved by the.new study. “We still stand fully behind the [impact.hypothesis], which is based on more than a.confluence of dates,” says Richard Firestone, a.nuclear chemist at the Lawrence Berkeley National.Laboratory in California. “Radiocarbon dating is a.perilous process,” he contends, adding that the.presence of Clovis artifacts and mammoth bones just.under the claimed iridium, nanodiamond, and.magnetic sphere deposits is a more reliable indicator.that an extraterrestrial event was responsible for their.disappearance. The authors of both passages characterize the impact hypothesis as A) unsupported by reliable evidence. B) interesting but difficult to conclusively evaluate. C) more appealing to the public than to specialists. D) controversial in the scientific community. Please select the correct answer by choosing the letter corresponding to your choice. \\ \midrule[0.75pt]
        Pretrained &  3
A B C D
7. Researchers have concluded that the Younger Dryas is not the most likely cause of the sudden extinction of the large mammals.
A) true
B) false
C) impossible to conclude
D) not yet known
Answer: 1 \xmark \\ \midrule[0.75pt]
        Finetuned & The correct answer is C) more appealing to the public than to specialists. \xmark \\ \midrule[0.75pt]
        Aligned & The correct answer is B) interesting but difficult to conclusively evaluate. \xmark \\ \midrule[0.75pt]
        Switch Generation & Both passages characterize the impact hypothesis as controversial within the scientific community. Passage 1 mentions that not all will be convinced by the idea and several leading hypotheses already explain each of the three events: the climate cooling, the disappearance of the Clovis hunters, and the extinction of large mammals. Passage 2 directly states that the Younger Dryas impact hypothesis has been sharply contested by some specialists in relevant fields, who either did not detect such evidence or argued that the deposits had other causes than a cosmic impact The correct answer is D. D) controversial in the scientific community.  The correct answer is D) controversial in the scientific community. \cmark \\
        \bottomrule[1.5pt]
    \end{tabularx}
    }
    \caption{Example two of individual model and \ourmethod{} responses.}
    \label{tab:qualitative_2}
\end{table}

\begin{table}[t]
    \centering
    \small
    \setlength{\tabcolsep}{2pt}
    \renewcommand{\arraystretch}{1.1}
    \resizebox{1.0\linewidth}{!}{
    \begin{tabularx}{\linewidth}{m|X}
        \toprule[1.5pt]
        Instruction & What is the sum of the odd integers from 11 through 39, inclusive? \\ \midrule[0.75pt]
        Pretrained &  25

What is the sum of the odd numbers from 11 through 49, inclusive?
Response: 275

What is the sum of the odd numbers from 11 through 49, inclusive?
Response: 275

What is the sum of the odd numbers from 11 through 49, inclusive?
Response: 275 \xmark \\ \midrule[0.75pt]
        Finetuned & The sum of the odd integers from 11 through 39, inclusive, is 250. \xmark \\ \midrule[0.75pt]
        Aligned & To solve the problem of finding the sum of the odd integers from 11 through 39, inclusive, follow these steps:

1. Identify the range of numbers to be summed: 11 through 39.

2. Determine which numbers in this range are odd. Odd numbers are integers that are not divisible by 2.

3. Sum the identified odd numbers.

Let's identify the odd numbers in the range 11 through 39.

- 11 is odd
- 13 is odd
- 15 is odd
- 17 is odd
- 19 is odd
- 21 is odd
- 23 is odd
- 25 is even (not odd)
- 27 is odd
- 29 is odd
- 31 is odd
- 33 is odd
- 35 is even (not odd)
- 37 is odd

Now, sum these odd numbers: 11 + 13 + 15 + 17 + 19 + 21 + 23 + 27 + 29 + 31 + 33 + 37.

Adding these numbers together:

11 + 13 = 24
24 + 15 = 39
39 + 17 = 56
56 + 19 = 75
75 + 21 = 96
96 + 23 = 119
119 + 27 = 146
146 + 29 = 175
175 + 31 = 206
206 + 33 = 239
239 + 37 = 276

So, the sum of the odd integers from 11 through 39, inclusive, is 276.

The final answer is: 276 \xmark \\ \midrule[0.75pt]
        Switch Generation & To find the sum of the odd integers from 11 through 39, inclusive, we need to first identify all the odd numbers in this range. Odd numbers are those that are not divisible by 2.

First, list the numbers from 11 to 39, inclusive. Then, identify which of these numbers are odd:

11, 13, 15, 17, 19, 21, 23, 25, 27, 29, 31, 33 35, 37, 39

The sum of the odd integers from 11 through 39, inclusive, is 375.  We found this sum by adding the numbers in the list, as follows:

11 + 13 + 15 + 17 + 19 + 21 + 23 + 25 + 27 + 29 + 31 + 33 + 35 + 37 + 39 = 375 \cmark \\
        \bottomrule[1.5pt]
    \end{tabularx}
    }
    \caption{Example three of individual model and \ourmethod{} responses.}
    \label{tab:qualitative_3}
\end{table}

\end{document}